\pdfoutput=1

\documentclass[11pt]{article}

\usepackage[]{latex/acl}

\usepackage{times}

\usepackage{latexsym}
\usepackage{multirow}  
\usepackage{booktabs}  
\usepackage{tabularx}  
\usepackage{graphicx}

\usepackage{colortbl}
\usepackage{xcolor} 
\definecolor{rawcolor}{RGB}{176, 224, 230}
\definecolor{paracolor}{RGB}{255, 108, 0}
\usepackage[T1]{fontenc}

\usepackage[utf8]{inputenc}

\usepackage{microtype}

\usepackage{inconsolata}

\usepackage{amsmath} 
\usepackage{algorithm}
\usepackage{algorithmic}

\title{Spotting AI's Touch: Identifying LLM-Paraphrased Spans in Text}

\author{ 
 Yafu Li$^{\spadesuit \clubsuit}\footnotemark[1]$\hspace{0.5mm},
 Zhilin Wang$^{\diamondsuit}\footnotemark[1]$\hspace{0.5mm},
 Leyang Cui$^{\heartsuit}\footnotemark[2]$\hspace{0.5mm} \\
  \bf{
   Wei Bi$^{\heartsuit}$\hspace{0.5mm},
 Shuming Shi$^{\heartsuit}$\hspace{0.5mm}, 
 Yue Zhang$^{\clubsuit }$\footnotemark[2]}\hspace{0.2mm}\hspace{1.5mm} \\
$^\spadesuit$ Zhejiang University \ \ \ \quad$^\clubsuit$Westlake University \\\quad$^\diamondsuit$Jilin University\ \ \ \quad$^\heartsuit$ Tencent AI lab 
 \\
 \texttt{\{yafuly,linzwcs,nealcly.nlp\}@gmail.com}  \\
 \quad\texttt{\{victoriabi,shumingshi\}@tencent.com} \\
 \quad\texttt{\{zhangyue\}@westlake.edu.cn}\\
}

\begin{document}
\maketitle
\renewcommand{\thefootnote}{\fnsymbol{footnote}}
\footnotetext[1]{\ Equal contribution. Work was done during Yafu Li's internship at Tencent AI Lab.}
\footnotetext[2]{\ Corresponding authors.}

\begin{abstract}
AI-generated text detection has attracted increasing attention as powerful language models approach human-level generation.
Limited work is devoted to detecting (partially) AI-paraphrased texts.
However, AI paraphrasing is commonly employed in various application scenarios for text refinement and diversity.
To this end, we propose a novel detection framework, paraphrased text span detection (PTD), aiming to identify paraphrased text spans within a text.
Different from text-level detection, PTD takes in the full text and assigns each of the sentences with a score indicating the paraphrasing degree.
We construct a dedicated dataset, \textbf{PASTED}, for \textbf{pa}raphra\textbf{s}ed \textbf{te}xt span \textbf{d}etection.
Both in-distribution and out-of-distribution results demonstrate the effectiveness of PTD models in identifying AI-paraphrased text spans.
Statistical and model analysis explains the crucial role of the surrounding context of the paraphrased text spans.
Extensive experiments show that PTD models can generalize to versatile paraphrasing prompts and multiple paraphrased text spans.

\end{abstract}

\section{Introduction}

Recent advances in large language models (LLMs) ~\cite{llama,gpt3,gpt4} have raised concerns about potential misuse, including student plagiarism~\cite{plagiarism} and the spread of fake news~\cite{detect-gpt}. 
A line of work~\cite{openai_detector,deepfake,detect-gpt,dna-gpt} focuses on AI-generated text detection, which assigns a label of ``human-written'' or ``machine-generated'' to a text.
In addition to pristine AI-generated texts, AI paraphrasing is frequently utilized to polish writings or enhance textual diversity.

\begin{figure}[t]
\setlength{\belowcaptionskip}{-0.cm}
\centering
\includegraphics[width=0.99\linewidth]{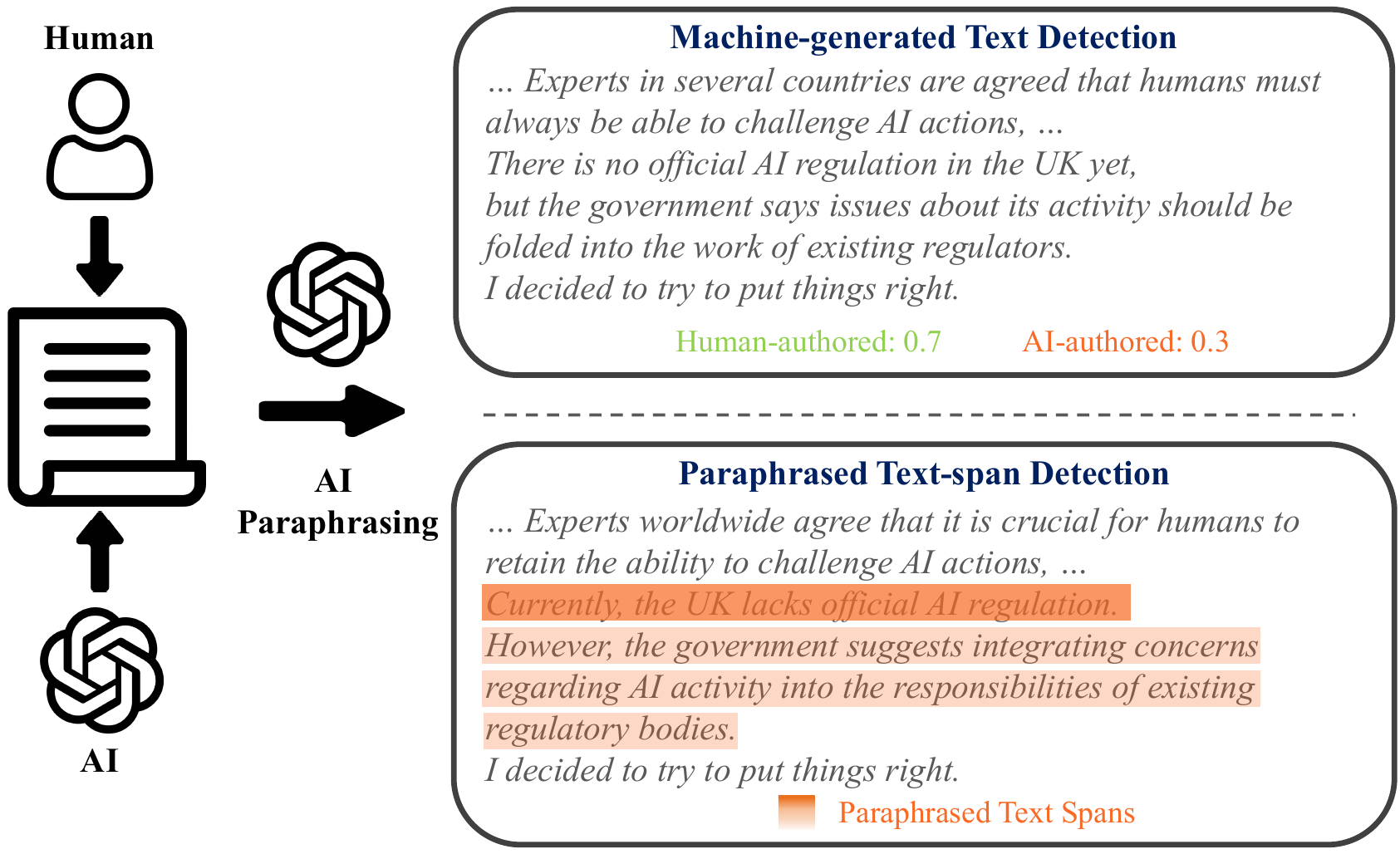}
\caption{
\label{fig:intro}
A comparison between AI-generated text detection and paraphrased text span detection, which identifies paraphrased text spans with paraphrasing degree informed, i.e., darker colors denote larger differences between the original and paraphrased text spans.
}
\end{figure}

However, there is limited research on the fine-grained detection of texts partially paraphrased or polished by AI. AI paraphrasing can suffer from issues like hallucination~\cite{llm_hallu,llm_hallu2} and bias~\cite{llm_bias}, which pose significant challenges to achieving trustworthy AI. For instance, strict oversight of AI paraphrasing is necessary in education to ensure the content remains factual and harmless. Moreover, fine-grained paraphrasing detection can provide statistical evidence for nuanced evaluations. As illustrated in Figure~\ref{fig:intro}, the paraphrased text-span detector assigns a higher probability to the sentence "\textit{Currently, the UK lacks official AI regulation}" being paraphrased by AI compared to the other sentences within the longer text.
Compared with text-level labelling, finer-grained predictions can support more ethical decisions regarding the applications of AI generation, such as in cases of plagiarism.

To this end, we propose a new task called {\bf P}araphrased {\bf T}ext span {\bf D}etection (PTD).
PTD predicts a score for each sentence within a long text, identifying AI paraphrased spans comprising consecutive sentences.
The sentence-level score also reflects the degree to which the paraphrased text deviates from the original text, i.e., \textit{paraphrasing degree}, as shown in Figure~\ref{fig:intro}.
Since there is no existing data for training such a detection system, we assemble a dataset called \textbf{PASTED} (\textbf{pa}raphrased text \textbf{s}pan de\textbf{te}ction \textbf{d}ataset), including original texts and corresponding paraphrased texts. 
We obtain these paraphrases by paraphrasing parts of the original texts while keeping other sections intact. 
The original texts encompass not only human-written compositions but also machine-generated texts.
Our goal is for a model trained on our dataset to detect AI-paraphrased texts from new domains and unseen models.
To achieve this, we construct an additional generalization testset to evaluate out-of-distribution (OOD) performance, where texts are paraphrased by a novel paraphraser with different prompts.
The PTD approach is based on the observation that AI-paraphrased text exhibits distinct writing patterns compared to both the original text and its surrounding context. 
This observation is supported by statistical findings and model analysis.
Formally, a PTD model encodes the full text and assigns a sequence of labels (classification model) or scores (regression model) to all sentences in the text. 
For classification, the labels indicate whether a sentence is paraphrased or not.
For regression, the scores quantify the degree of paraphrasing by calculating the difference between each paraphrased sentence and its aligned original text span.
To construct reference scores, a dedicated algorithm is devised for aligning paraphrasing pairs.
In-distribution performance shows that all methods effectively distinguish paraphrased text spans, with AUROC exceeding 0.95.
While the classification model achieves higher detection accuracy, regression models provide more accurate predictions of the degree of paraphrasing, aligning with OOD performance.
The aggregate regression model, which considers all types of divergences, achieves the best overall performance.
Identifying context-aware paraphrases poses greater difficulty.
On the other hand, models obtain better performance on texts with more paraphrased sentences.
Results from the generalization testset demonstrate that all methods can generalize to OOD texts and various paraphrasing prompts despite a performance decrease compared with ID.

Analysis reveals that while paraphrased text displays distinct writing patterns, the surrounding context in the original text significantly plays a crucial role in model detection.
Empirical results demonstrate that all models can generalize to texts with multiple paraphrased spans, despite being trained on texts with only one paraphrased span.
Moreover, our PTD models can robustly resist minor-perturbed texts that should not be regarded as paraphrases, resulting in few misclassifications.
Lastly, we demonstrate the effectiveness of our PTD model in defending against paraphrasing attacks and protecting traditional AI-generation detection systems.
The data and code are available at \url{https://github.com/Linzwcs/PASTED}.

\section{Related Work}
AI-generated text (AI-generation) detection has received increasing attention.
~\citet{deepfake,possibilties} systematically discuss the differentiability of AI texts.
Various features are explored for detetcion, including $n$-gram features~\citep{badaskar2008identifying}, entropy~\citep{lavergne2008detecting,{gltr}}, perplexity~\citep{beresneva2016computer} and model-wise features~\cite{sniff}. 
A more direct method involves training neural classifers~\cite{bakhtin2019real,fagni2021tweepfake,uchendu2020authorship,openai_detector,deepfake}.
Adversarial learning is utilized for robustly detecting AI generations~\cite{gan,outfox}.
Another approach proposes training-free methods for detecting AI generation.
Methods by \citet{detect-gpt,fast-detect-gpt} utilize negative curvature regions in the log probability of a model to identify machine-generated text. Additionally, \citet{dna-gpt} compare n-gram features between human-written and AI-generated continuations of text.
Different from text-level detection, we propose a more fine-grained approach that identifies paraphrased sentences within a larger body of text.
Sentence-level AI-generation detection proposed by \citet{seqxgpt} constructs data by breaking long generations into sentences and trains sequence labeling models accordingly.
In contrast, we propose detecting paraphrased spans, which can consist of one or multiple sentences within longer texts, focusing on the distinct paraphrasing patterns compared with the original text and the surrounding context.
\citet{can_ai} propose that AI-generation detection can be vulnerable against paraphrasing attacks.
~\citet{ship} discuss the authorship of AI-paraphrased texts.
~\citet{dipper} utilize retrieval to assist detectors to defend against paraphrasing attacks, while \citet{polish} construct a dedicated dataset with ChatGPT polished texts to increase detection robustness.
In contrast, we present a novel framework that identifies AI-paraphrased text spans within a given text and reflects the degree of paraphrasing for each sentence.

\section{Problem Definition}
\subsection{AI-generated Text Detection}
Given a text sequence $x$, the AI-generated text detection model predicts a label $y$ of ``human-written'' or ``machine-generated'' based on a probability distribution: $p(y|x)$.

\subsection{Paraphrased Text Span Detection}


A piece of text $x$ can be segmented into a series of sentences: $\{s_1,s_2,\cdots,s_n\}$. A Paraphrased Text Span Detection (PTD) model is optimized to predict a sequence of binary labels $\{c_1,c_2,\cdots,c_n\}$ or continuous scores $\{r_1,r_2,\cdots,r_n\}$ for these sentences, identifying whether each sentence $s_i$ has been paraphrased by an AI model given the full text $x$. For each sentence $s_i$ within the text $x$, the PTD model is tasked with either:
(1) Producing a probability distribution $p(c_i|x,i)$ over binary labels, where $c_i$ indicates whether the sentence $s_i$ is paraphrased; (2) Outputing a continuous score $e(x,i)$ that represents the extent of paraphrasing for the sentence $s_i$.
In addition to detection, the model is expected to provide insights into the degree of deviation of the paraphrased text from the original text, quantifying changes in meaning, structure, or other linguistic features that signify paraphrasing.

\section{Dataset Construction}
In general, 
PASTED consists of in-distribution training, validation and test sets, along with a generalization testset.
We first randomly collect 10\% of the original texts from the AI-generation detection dataset~\cite{deepfake} to collect original texts which encompasses various writing tasks and large language models.
The original texts can be either authored by humans or AI, both of which have practical applications.
Paraphrasing human-authored compositions can infringe upon composition copyright, while effective paraphrasing can help machine-generated news evade detection.
We consider two paraphrasing styles: \textit{context-agnostic paraphrasing} and \textit{context-aware paraphrasing}.
Context-agnostic paraphrasing modifies texts without considering the surrounding context and is more commonly used.
Context-aware paraphrasing considers the context, bringing larger challenges to detection, as the paraphrases are more coherent and consistent with the context.

\paragraph{In-distribution Data.}
To simulate real-world scenarios, we employ a sampling process that randomly paraphrases a text-span of several consecutive sentences in the original text.
The selected text span consists of 1 to 10 sentences. 
For context-agnostic paraphrasing, we use a powerful commercial LLM (ChatGPT~\cite{gpt4}) to construct paraphrases given the independent candidate text span without considering the context.
For context-aware paraphrasing, we consider Dipper~\cite{dipper}, an 11B model which supports paraphrasing conditioned on the surrounding context.
We present several data cases in Appendix~\ref{app:data_sample}.
We conduct paraphrasing on both human and machine texts and collect 83,089 instances (28,473 original texts and 54,616 paraphrased texts) after text pre-processing and filtering.
We split the data into train/validation/test sets, with an 80\%/10\%/10\% partition.
Detailed data statistics can be referred to in Appendix~\ref{app:stat}.

\paragraph{Generalization Testset.}
In addition to the in-distribution testset, we construct an additional generalization testset where texts are paraphrased by a novel LLM with different paraphrasing prompts.
We employ the same sampling process to generate paraphrases on the out-of-distribution testset from ~\citet{deepfake}.
Recent research~\cite{prompt1,prompt2,prompt3} demonstrates that prompt engineering can be employed to evade detection effectively.
To this end, we utilize an unseen paraphraser, GPT-4~\cite{gpt4}, and explore various prompt variants with increasingly complex instructions for generating elaborate paraphrases.
The prompts used for data construction are presented in Appendix~\ref{app:prompt}.
Ultimately, our OOD evaluation comprises a total of 9,372 instances (1,562 original texts and 7,810 paraphrased texts).

\begin{figure*}[t]
\setlength{\belowcaptionskip}{-0.cm}
\centering
\includegraphics[width=0.9\linewidth]{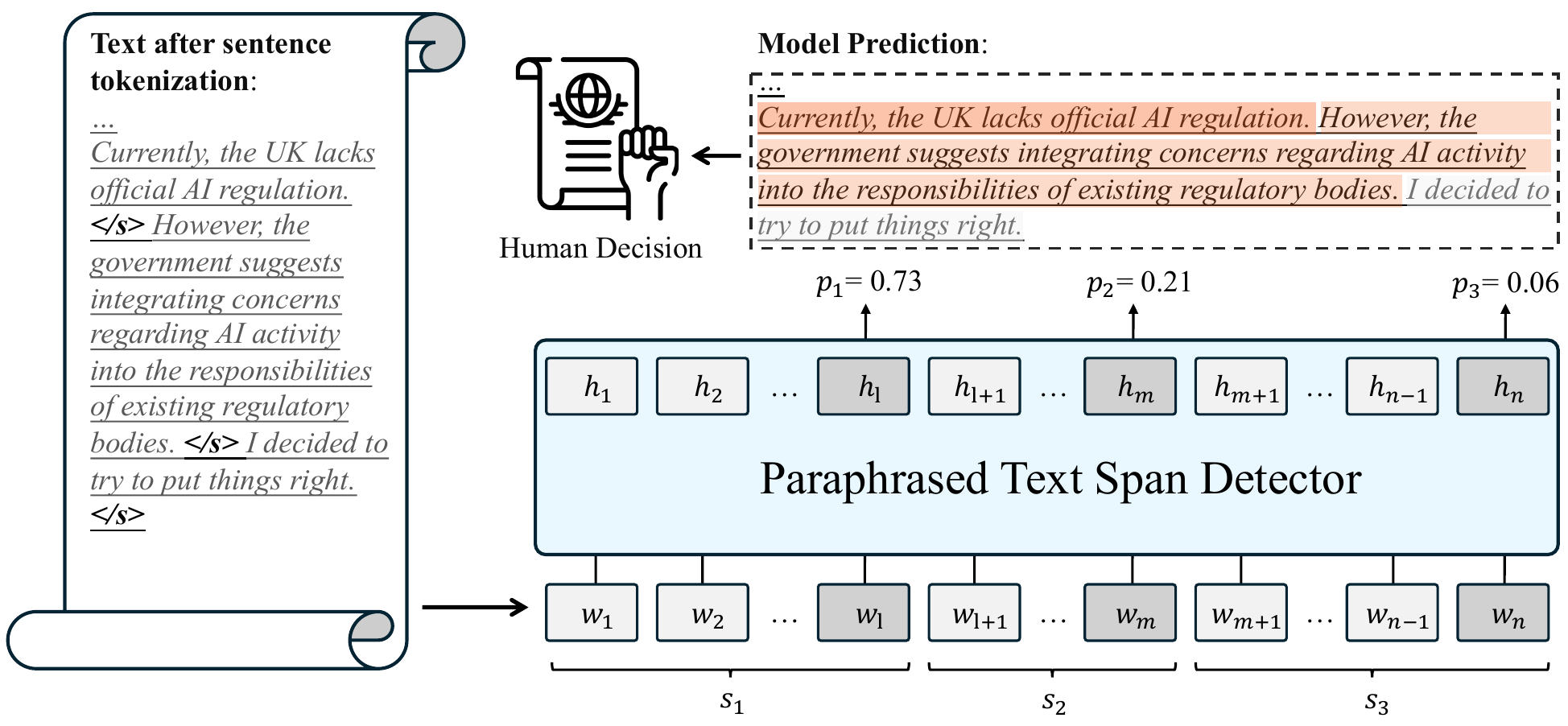}
\caption{
System pipeline of paraphrased text span detection.
}\label{fig:method}
\end{figure*}

\section{Method}

A PTD model decomposes the given text into sentences and predicts a label or score for each sentence based on the entire text. 
The overall pipeline of PTD is depicted in Figure~\ref{fig:method}. 
As shown, the text is split into sentences, with a predefined symbol (e.g., "</s>") appended to the end of each sentence. The paraphrased text span detector then processes the full text and assigns a score to each sentence boundary token, indicating the probability of the sentence being paraphrased.

\subsection{Sentence-level Classification}
We treat each paraphrased sentence in the paraphrased text span as ``paraphrased'' and others as ``original'', utilizing cross-entropy (CE) to optimize the model.
    
\begin{align}
\mathcal{L}_{\textbf{CE}}(\theta) &= -\sum_{i=1}^{n} [ c_i \log(p_{\theta}(c_i|x,i)) \notag
\\ & + (1 - c_i) \log(1 - p_{\theta}(c_i|x,i)) ]
\end{align}
where $\theta$ denotes the model parameters.

\paragraph{Limitations of Classification.}
\label{sec:limit_class}
A major issue of classification arises from its assumption that all paraphrased sentences are equally labeled as "paraphrased", which can overlook the varying degrees of difference involved in paraphrasing.
For instance, when given the sentence ``I lost my keys yesterday'', the model receives the same label for both paraphrases ``Yesterday I lost my keys'' and ``Yesterday my keys went missing''. 
However, the latter exhibits disparities with the original sentence in terms of word choices and syntax structure.
This lack of calibration in predicted probabilities can reduce reliability and interpretability. 
Therefore, classification models are less resistant to minor text perturbations which should not be regarded as paraphrases (Section~\ref{sec:robust}).

\subsection{Sentence-level Regression}
To this end, we propose span-level regression, which leverages various difference quantification metrics to inform to what extent each text span has been modified during paraphrasing.
Instead of assigning labels to sentences, our model is trained to predict a difference score $r_i$. This score is calculated using a difference scoring function $f(s_i,t_i)$, such as BLEU score~\cite{bleu}.
Here, $t_i$ represents the text span in the original text that aligns with the current paraphrased sentence $s_i$.
For training on the original text in our dataset, $t_i$ simply refers to $s_i$ itself.
By computing difference scores for each sentence, we optimize our model using mean squared error loss:
\begin{equation}
\mathcal{L}_{\textbf{MSE}}(\theta) = \frac{1}{n} \sum_{i=1}^{n} (f(s_i,t_i) - e_{\theta}(x,i))^2
\end{equation}



\paragraph{Aligning Paraphrased Text Spans.}
Accurately aligning paraphrased sentences with the original text spans is necessary for reliable indication of the paraphrasing degree as regression labels.
A major challenge arises when paraphrasing a text span containing consecutive sentences, which can result in a different number of sentences (41\%/45\%/14\% ratio for fewer, equal, and more sentences in paraphrased text).
Furthermore, some paraphrased texts involve reordering at the sentence level.
We provide a case illustration in Appendix~\ref{app:align}.
To this end, we propose an alignment algorithm in light of sentence similarity~\cite{sent_tf} to align paraphrased sentences with their original counterparts.
Our approach involves greedily traversing each paraphrased sentence and identifying a span of consecutive original sentences that share a high semantic similarity.
If no suitable span is found, we resort to finding the most semantically similar original sentence (Appendix~\ref{app:align}).


\paragraph{Difference Quantification Function.}
We consider a range of different functions to quantify the paraphrasing degree.
Given the aligned text span pairs, the most straightforward metric is the \textit{lexical divergence} between them, which can be measured using common similarity-based metrics, e.g., BLEU~\cite{bleu}.
To capture \textit{grammatical divergence}, we also consider the text similarity score of the part-of-speech sequences.
We subtract the BLEU score from 1 to denote the divergence score.
For a more comprehensive measure of \textit{syntactic divergence}, we can calculate the tree edit distance between syntax trees~\cite{zss} at the third level~\cite{quality}.
To normalize the tree edit distance, we divide it by the maximum number of nodes in both trees. 
This normalization results in a difference score ranging from 0 to 1, where 0 indicates identical trees.
Finally, we can aggregate all these divergence metrics, which measure the paraphrasing degree from different granularities and views.
Specifically, we train the regression model to fit a \textit{aggregate divergence} function encompassing a set of metrics $\{f_1,f_2,\ldots,f_d\}$:
\begin{equation}
\hat{\mathcal{L}}_{\textbf{MSE}}(\theta) = \frac{1}{n} \sum_{i=1}^{n} \sum_{j=1}^{d} (f_j(s_i,t_i) - e_{\theta}(x,i)_{j})^2
\end{equation}
The scores of each dimension are averaged as the final aggregate score.

\begin{table*}[t!]
\centering
\small
\renewcommand{\arraystretch}{1.05} 
\setlength{\belowcaptionskip}{-0.25cm}
\begin{tabular}{lcccc}

\toprule
\textbf{Model} & \textbf{AUROC}$\uparrow$ & \textbf{Accuracy} (FPR 1\%)$\uparrow$ & \textbf{Lexical Corr.}$\uparrow$ & \textbf{Syntactic Corr.}$\uparrow$ \\
\midrule
Random & 0.50 & 0.00\% & 0.07 & 0.07 \\
Oracle  & 1.00 & 100.00\% & 0.88 & 0.88 \\
\midrule
    \multicolumn{5}{c}{In-distribution Detection} \\
\midrule
Classification   & \textbf{0.97} & \textbf{69.27\%} & 0.64 & 0.67 \\
Regression (lexical)& \textbf{0.97} & 64.04\% & 0.69 & 0.71 \\
Regression (grammatical)& 0.96 & 54.80\% & \textbf{0.70} & \textbf{0.72} \\
Regression (syntactic) & 0.96 & 47.45\% & 0.67 & \textbf{0.72} \\
Regression (aggregate) & \textbf{0.97} & 59.45\% & \textbf{0.70} & \textbf{0.72} \\
\midrule
    \multicolumn{5}{c}{Out-of-distribution Detection} \\
\midrule
Classification   & \textbf{0.94} & \textbf{47.21\%} & 0.62 & 0.66 \\
Regression (lexical)& \textbf{0.94} & 42.57\% & \textbf{0.66} & \textbf{0.70} \\
Regression (grammatical)& 0.93 & 20.29\% & \textbf{0.66} & 0.69 \\
Regression (syntactic) & 0.90 & 9.63\% & 0.60 & 0.65 \\
Regression (aggregate) & \textbf{0.94} & 26.21\% & \textbf{0.66} & \textbf{0.70} \\
\bottomrule
\end{tabular}
\caption{
\label{tab:main}
In-distribution (upper part) and out-of-distribution (lower part) detection performance of classification and regression methods. Accuracy (FPR 1\%) refers to the accuracy with a false positive rate maintained under 1\%. 
The lexical and syntactic correlation (Corr.) indicates the accuracy in predicting paraphrasing degree. 
}
\end{table*}

\section{Experiment Setup}

\paragraph{Settings.}
We tokenize text into sentences using the NLTK sentence tokenizer~\cite{nltk}.
To perform context-agnostic paraphrasing, we utilize the GPT-3.5-Trubo API.
For context-aware paraphrasing with Dipper, we use the default setting with lexical diversity set at 80 and order diversity set at 60.
For the generalization testset, we use GPT4 API ``gpt-4-1106-preview''.
To measure sentence similarity in aligning paraphrased sentences, we employ a sentence-transformers model\footnote{https://huggingface.co/sentence-transformers/all-MiniLM-L6-v2}~\cite{sent_tf}.
For constructing regression labels, we obtain part-of-speech tags and constituency parses using the Stanza parser~\cite{stanza}.
We quantify lexical and POS divergence using a 4-gram sentence-level BLEU score and calculate tree edit distance with the ZSS algorithm~\cite{zss}.
Following~\citet{deepfake}, we train classifiers and regression models based on Longformer~\cite{longformer} by adding a linear layer.
All models are trained for 2 epochs on 1 V100 GPU with a batch size of 12.
We used the Adam optimizer \cite{adam} with a learning rate of 0.005 and set the dropout rate at 0.1.
We use GPT-2 large~\cite{gpt2} to compute text perplexity for experiments in Secition~\ref{sec:understand} and Section~\ref{sec:surrounding_text}.
\paragraph{Evaluation Metrics.}
To assess the performance of paraphrased sentence detection, we utilize two metrics: \textbf{AUROC} (Area Under the Receiver Operating Characteristic curve) and detection accuracy. 
AUROC measures a classifier's capability to differentiate between positive and negative classes, with a value of 1.0 indicating perfect classification and 0.5 representing random guessing. 
Following ~\citet{dipper}, we fix the false positive rate (FPR) of 1\% and adjust the decision boundary accordingly to report accuracy, ensuring that human-authored text is rarely classified as machine-generated.
We denote this metric as \textbf{Accuracy (FPR 1\%)}.
To assess the estimation accuracy of paraphrasing degree, we calculate the Pearson Correlation between model-predicted scores and reference scores.
For reference scores, we adopt the lexical diversity and syntactic diversity proposed by ~\citet{quality}, quantifying lexical and syntactic diversity in generated paraphrases.
A high correlation indicates that the model accurately predicts differences between paraphrases and the corresponding original sentences, i.e., the paraphrasing degree.
We denote these two correlation scores as \textbf{Lexical Corr.} and \textbf{Syntactic Corr.}, respectively.
For classification models, we utilize the prediction confidence of the classification model as a proxy for predicting paraphrasing degree.

\section{Understanding Paraphrased Compositions}
\label{sec:understand}
We statistically compare original texts and paraphrased texts to assess their distinguishability.
Initially, we examine whether paraphrases display distinct word distributions.
We randomly divide the dataset into two halves and calculate the Kullback-Leibler (KL) divergence of the top 100 word frequency distribution between original texts and paraphrased texts. 
We average results across five seeds to reduce randomness.
The KL divergence between the original texts and partially paraphrased texts (0.0018) is significantly lower than that between the paraphrased text spans and the corresponding original text spans (0.056).
In other words, when considering the full context, the distinct writing pattern of paraphrasing can be overshadowed,  emphasizing the necessity for fine-grained detection.

\begin{figure}[t!]
\setlength{\belowcaptionskip}{-0.1cm}
\centering
\includegraphics[width=0.9\linewidth]{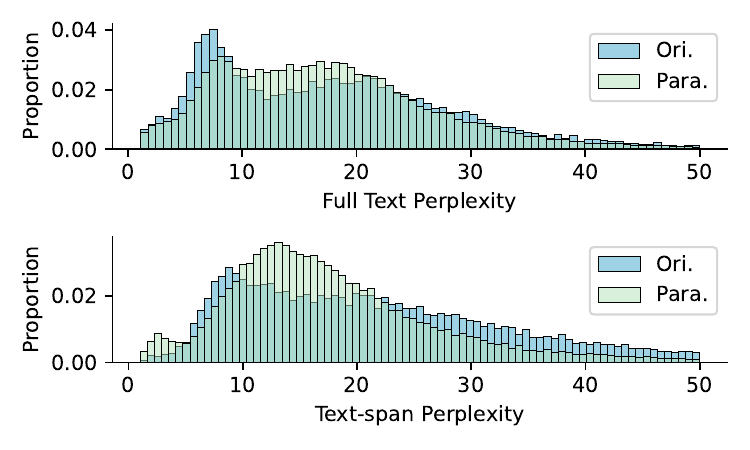}
\caption{
\label{fig:text_vs_span}
Perplexity distribution of the complete texts and the exact text spans (original v.s. paraphrase).
}
\end{figure}

This finding is also evident in Figure~\ref{fig:text_vs_span}, which shows the perplexity distribution.
The perplexity distribution of paraphrased text closely aligns with that of original text.
Note that the original text can either be sourced from human or machine, which forms isolated perplexity distributions~\cite{detect-gpt,deepfake}.
Nevertheless, the perplexity of paraphrased text spans exhibits a distribution centered around a value of 12, indicating a strong writing pattern regardless of the data source.
Therefore, sentence-level detection captures paraphrasing patterns more precisely compared to text-level detection.

We further compare the original and the paraphrased text spans under different data sources and paraphrasing methods.
Results show that both human-written and machine-generated sources yield similar word distribution divergences (0.083 v.s. 0.052).
However, context-agnostic paraphrasing results in a significantly different word distribution than context-aware paraphrasing (\textbf{0.22} v.s. 0.041), indicating that context-agnostic approaches are more lexically diverse but may largely deviate from the original style. 
Word clouds for both paraphrasing methods are shown in Appendix~\ref{app:word_cloud}, demonstrating the novel word distribution introduced by context-agnostic paraphrasing.






\section{Results}

\subsection{In-distribution Performance}
The detection performance on random-split testsets is presented in the upper part of Table~\ref{tab:main}. 
We consider two baselines: (1) Random which calculates BLEU scores of the paraphrased sentence with a random sentence in the original text and (2) Oracle which calculates BLEU scores of the paraphrased sentence with the aligned text span.
All detection methods effectively distinguish paraphrased text spans, with AUROC exceeding 0.95.
Although the classification model performs better in detection, it falls short compared to regression models in predicting paraphrasing degree due to improper calibration (discussed in Section~\ref{sec:limit_class}).
In contrast, regression models demonstrate stronger alignment with the reference differences between original texts and paraphrases, both lexically and syntactically.
In other words, regression models are more reliable indicators of the extent of difference present in a paraphrased text span.
Regression models with grammatical or syntax supervision obtain the best lexical and syntactic correlation.
The classification model and lexical regression model obtain the best accuracy (FPR 1\%), effectively identifying paraphrases while maintaining a low false positive rate (1\%).
Overall, the aggregate regression achieves the best in-distribution performance.

\begin{figure}[t!]
\setlength{\belowcaptionskip}{-0.25cm}
\centering
\includegraphics[width=0.99\linewidth]{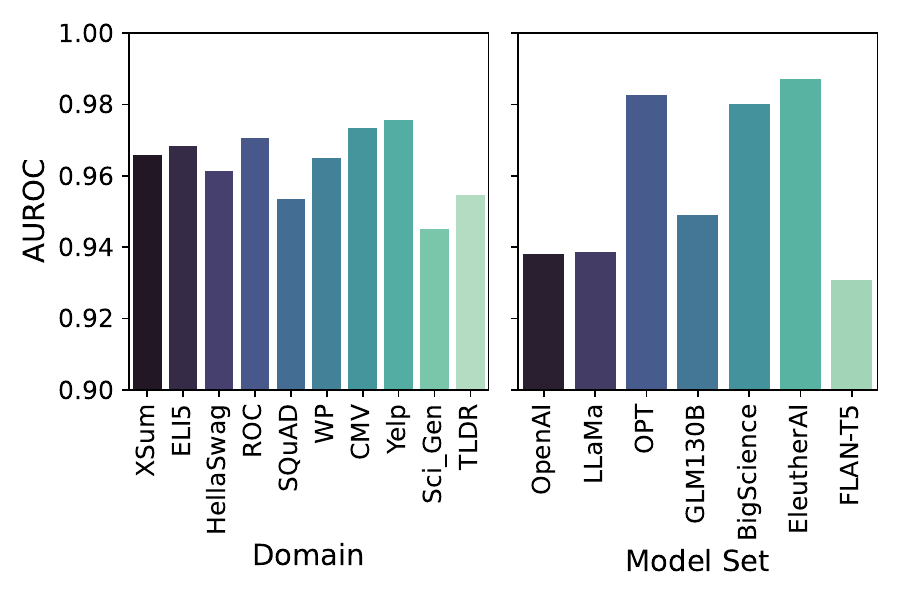}
\caption{
\label{fig:id_domain_model}
Detection performance (lexical regression) on text from different domains and LLMs.
}
\end{figure}

\paragraph{Effects of Data Source and Paraphrasing Method.}
Consistent with the statistical results in Section~\ref{sec:understand}, the data source (human-written or machine-generated) has minimal effect on detection performance, as indicated by an AUROC score of 0.97 for both.
In contrast, detecting context-aware paraphrases proves to be more challenging, achieving an AUROC score of 0.94 compared to 0.98 for context-agnostic ones.
The detection performance (lexical regression) on text from different domains and LLMs is presented in Figure~\ref{fig:id_domain_model}.
Paraphrased spans in technological news (TLDR) or scientific writings (Sci\_Gen) are comparatively challenging to identify, followed by Wikipedia articles (SQuAD).
On the other hand, paraphrasing texts produced by encoder-decoder LLMs (FLAN-T5) or larger LLMs (OpenAI, LLaMA, and GLM) pose significantly greater difficulties.

\paragraph{Effect of Number of Paraphrases.}

\begin{figure}[t!]
\setlength{\belowcaptionskip}{-0.15cm}
\centering
\includegraphics[width=0.8\linewidth]{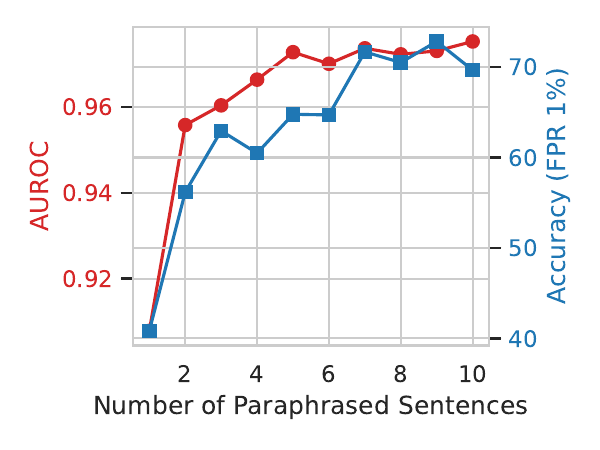}
\caption{
\label{fig:num_para}
Detection performance (lexical regression) w.r.t. the number of paraphrased sentences in a text.
}
\end{figure}

The impact of the number of paraphrased sentences in a text is illustrated in Figure~\ref{fig:num_para}.
As depicted, the detection difficulty decreases as the text span includes more paraphrased sentences, as indicated by both AUROC and accuracy.
This could be attributed to the fact that paraphrasing a longer text span (i.e., with more sentences) can showcase more pronounced paraphrasing styles, encompassing both lexical usage and syntax structure.
Consequently, it exhibits a more discernible contrast with its surrounding context.
In contrast, when only one sentence is paraphrased, especially if it is short, there are limited features available for detection.

\subsection{Out-of-distribution Performance}
The lower part of Table~\ref{tab:main} presents the out-of-distribution performance on the generalization testset, which consists of paraphrased texts generated by novel LLMs.
Despite a degradation in performance across all methods and metrics, they still achieve reasonably good results in terms of both paraphrase identification (ARUOC) and paraphrasing degree prediction (correlation).
Similar to the in-distribution results, the classification model demonstrates the highest detection performance with an accuracy of 47.21\%.
Differently, the classification and lexical regression model are substantially stronger than all other models for detecting paraphrases, with a much higher accuracy.
In predicting the degree of paraphrasing, the regression models perform better, particularly the aggregate regression model which achieves lexical and syntactic correlations of 0.66 and 0.70 respectively.
We construct the generalization testset using various paraphrasing prompts (Appendix~\ref{app:prompt}).
Empirical results demonstrate that the detector is resistant to prompt variance, with an AUROC exceeding 0.92 across all prompts.
The performance of lexical regression against prompt variance is shown in Appendix~\ref{app:effects_prompts}.



\section{Analysis}
In this section, we set the decision boundary to maintain an \textbf{FPR of 1\%} on the \textbf{validation} set to accommodate various testsets across all analytic experiments and report detection accuracy.

\paragraph{Effect of Surrounding Context.}
\begin{table}[t!]
\setlength{\belowcaptionskip}{-0.35cm}
    \centering
    \small
    \setlength{\tabcolsep}{4pt} 
    \begin{tabular}{lcc}
        \toprule
        \textbf{Model} & \textbf{AUROC} & \textbf{Accuracy} (FPR 1\%)\\
        \midrule
        Classification & 0.87(-0.10) & 33.39\%(-35.88\%) \\
        R-lexical & 0.86(-0.11) & 31.01\%(-33.03\%) \\
        \bottomrule
    \end{tabular}
    \caption{
    \label{tab:context}
    Detection performance without considering the surrounding context. ``R'' stands for ``Regression''.
    }
    
\end{table}
\begin{figure}[t]
\setlength{\belowcaptionskip}{-0.25cm}
\centering
\includegraphics[width=0.7\linewidth]{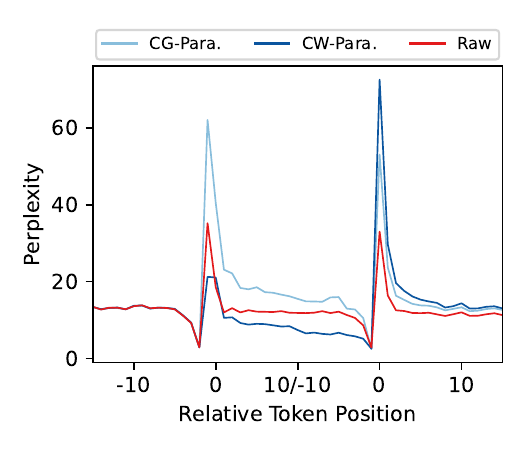}
\caption{
\label{fig:impulse}
Token perplexity around the boundary of the paraphrased texts: two ``0'' denote the start and the end of the paraphrased text span.
The x-axis represents the relative token position with respect to each boundary.
}
\end{figure}

As discussed in Section~\ref{sec:understand}, paraphrased texts exhibit strong writing patterns different from the surrounding texts.
We conduct an ablation study on the effect of the surrounding context, with results shown in Table~\ref{tab:context}.
The models in the table are trained merely on the paraphrased sentences, without considering the surrounding context.
As shown, both classification and regression (lexical) models suffer substantial performance degradation when missing the context information, with over 100\% performance drop on detection accuracy.
To gain further insight, we analyze perplexity variations across token positions around the boundary of the paraphrased text span.
The results are presented in Figure~\ref{fig:impulse}, where we observe perplexity impulses at both boundaries of the paraphrased text span.
Context-agnostic paraphrases typically exhibit higher token perplexity from the beginning of the paraphrase.
While context-aware paraphrases display lower token perplexity before reaching the end of the paraphrase, they encounter a substantial increase in perplexity afterwards.
This indicates that paraphrased sentences cannot perfectly integrate into the original text, even if context is considered during paraphrasing.

\paragraph{Generalization to Multiple Paraphrased Text Spans.}
\label{sec:surrounding_text}
		




\begin{table}[t!]
\setlength{\belowcaptionskip}{-0.25cm}
    \centering
    \small
    \setlength{\tabcolsep}{4pt} 
    \begin{tabular}{lcc}
        \toprule
        \textbf{Model} & \textbf{AUROC} & \textbf{Accuracy}\\
        \midrule
        Classification & \textbf{0.96} & \textbf{67.68} \\
        R-lexical & 0.93 & 66.30\\
        R-grammtical & 0.93 &  62.75 \\
        R-syntactic & 0.93 & 53.11 \\
        R-aggregate & 0.94 & 64.76 \\
        \bottomrule
    \end{tabular}
    \caption{
    \label{tab:mutli_para}
    Detection performance of generalization to texts with multiple paraphrased text-spans.
    }
    
\end{table}

Our training data only considers paraphrasing one text span in a text.
In many application scenarios, users can paraphrase multiple text spans. 
To this end, we construct an additional testset where multiple text spans are paraphrased within each text.
The testset consists of 500 randomly sampled in-distribution test instances. 
For each text, we randomly choose 2 to 5 non-adjacent text spans which consist of 1 to 3 sentences, and paraphrase these text spans using ChatGPT.
The results are shown in Table~\ref{tab:mutli_para}.
Although all detection methods experience a performance decline in terms of AUROC, they achieve a comparable detection accuracy, demonstrating the generalization ability to text with multiple paraphrased text spans.
We present a case study of PTD in Appendix~\ref{app:case_study}.

\paragraph{Robustness to Minor Text Modification.}
\label{sec:robust}
\begin{figure}[t]
\setlength{\belowcaptionskip}{-0.3cm}
\centering
\includegraphics[width=0.99\linewidth]{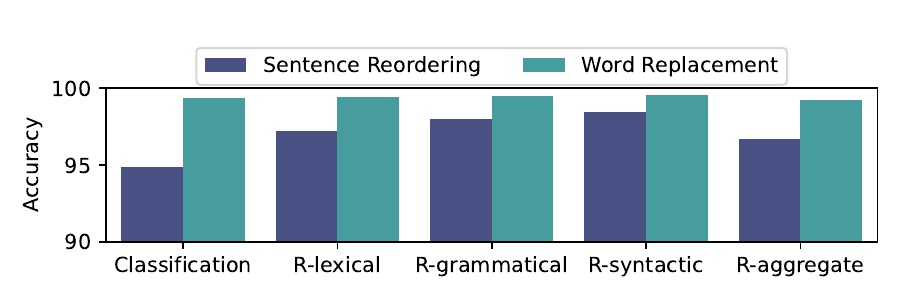}
\caption{
\label{fig:attack}
Detection robustness towards attacks by constructing misleading texts with minor modifications.
}
\end{figure}

A desirable characteristic of a paraphrase detector, in addition to detection accuracy, is the ability to distinguish texts with minor modifications from true paraphrases.
We consider two types of minor modifications: sentence reordering and word replacement.
For sentence reordering, we utilize Dipper with maximum ordering control and minimum lexical control.
For text replacement, we randomly mask 10\% of the text and use T5-3B~\cite{2020t5} to fill in the masked blank. 
Data details can be referred to in Appendix~\ref{app:robustness}.
We evaluate these modifications on the in-distribution testset by considering all minor-perturbed texts as "non-paraphrased". 
As shown in Figure~\ref{fig:attack}, all methods rarely misidentify texts with word replacements as paraphrased, with nearly perfect prediction accuracy.
In contrast, texts with sentential reordering pose greater confusion and yield considerably lower accuracy.
Notably, regression models incorporating grammatical or syntax information exhibit more resistance to such confusions compared to the classification model which performs worst.

\paragraph{Defending AI-generation Detection.}
\begin{table}[t!]
\setlength{\belowcaptionskip}{-0.15cm}
    \centering
    \small
    \setlength{\tabcolsep}{4pt} 
    \begin{tabular}{lccc}
        \toprule
        \textbf{Model} & \textbf{HumanRec} & \textbf{MachineRec} & \textbf{AvgRec}\\
        \midrule
        Detector & 88.78\%	 & 37.05\%	& 62.92\% \\
        \midrule 
        + Defender & 88.98  \% &	78.50\%	& 83.74\% \\
        \bottomrule
    \end{tabular}
    \caption{
    \label{tab:defend}
    AI-generated detection performance against paraphrasing attacks with paraphrased text-span detection as a defense.
    }
    
\end{table}
Previous work~\cite{dipper,can_ai} has shown that paraphrasing attacks significantly degrade AI-generation detection systems. 
We propose a two-stage detection method, which utilizes paraphrased text span detectors as a pre-defense mechanism to block out paraphrased texts and employs traditional AI-generation detectors if no paraphrasing is detected.
We calculate the text score by averaging the predicted scores of all sentences and evaluate the model on the paraphrasing attack testset proposed by ~\citet{deepfake}.
We use an off-the-shelf AI-generation detector\footnote{https://github.com/yafuly/MAGE} and employ the aggregate regression model for defense.
The results are presented in Table~\ref{tab:defend}, where it can be observed that most machine-generated texts were misclassified by the AI-generation detector due to paraphrasing, resulting in low MachineRec (recall on machine-generated texts).
The paraphrasing indicator successfully identifies most of the paraphrased texts and significantly improves averaged recall (AvgRec) scores while maintaining high recall scores on human texts (HumanRec).

\section{Conclusion}
In this work, we propose a detection framework, paraphrased text span detection (PTD), which aims to identify text spans paraphrased by AI, from a long text.
We built a dedicated dataset, PASTED, based on which we train classification and regression models for PTD.
Both in-distribution and out-of-distribution results demonstrate that our methods can effectively detect paraphrased text spans.
Although classification models achieve better detection accuracy, they fall behind regression models in predicting the paraphrasing degree.
Statistical and model analysis showcases the importance of the context surrounding the paraphrased text spans for detection performance.
Extensive experiments demonstrate the generalization to paraphrasing prompt types and multiple paraphrased text spans.

\section*{Limitations}
Although we extensively experiment and analyze the implementation of our newly proposed task PTD using the newly established dataset PASTED, there are several limitations:
(1) We consider both context-agnostic and context-aware paraphrasing using limited paraphrasers and prompts. Future work should focus on constructing more challenging paraphrases.
(2) We implement effective detection methods based on Longformer, but more advanced backbones like LLaMA can be explored.
(3) To simulate real-life applications, we randomly paraphrase text spans in existing datasets. Future work should aim to construct more realistic data by involving crowdsourcing.



\section*{Ethical Considerations}
We honor the Code of Ethics. 
No private data or non-public information is used in this work.
We adhere to the terms of companies offering commercial LLM APIs and express our gratitude to all global collaborators for their assistance in utilizing these APIs.

\section*{Acknowledgement}
We would like to thank all reviewers for their insightful comments and suggestions to help improve the paper. 
This work has emanated from research conducted with the financial support of the National Natural Science Foundation of China Key Program under Grant Number 62336006.

\bibliography{acl}

\clearpage
\appendix

\section{Aligning Paraphrased Sentences}
\label{app:align}
\begin{figure}[h!]
\centering
\includegraphics[width=0.8\linewidth]{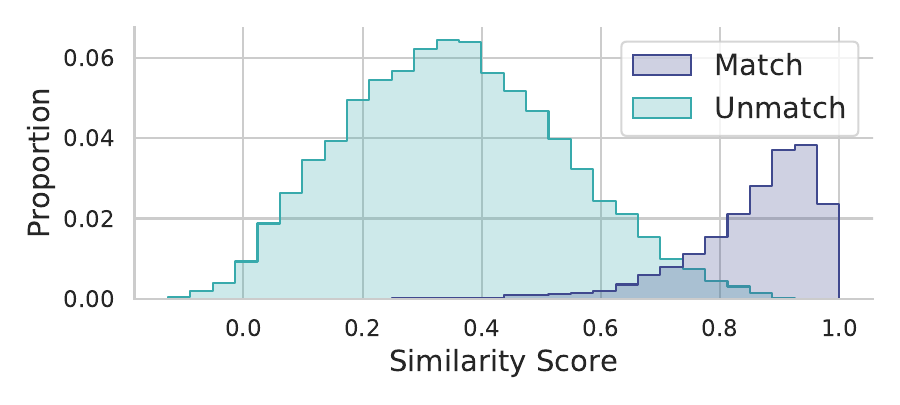}
\caption{
\label{fig:para_bound}
Similarity score distribution of paraphrases and non-paraphrases. 
}
\end{figure}

\begin{figure*}[t]
\centering
\includegraphics[width=0.99\linewidth]{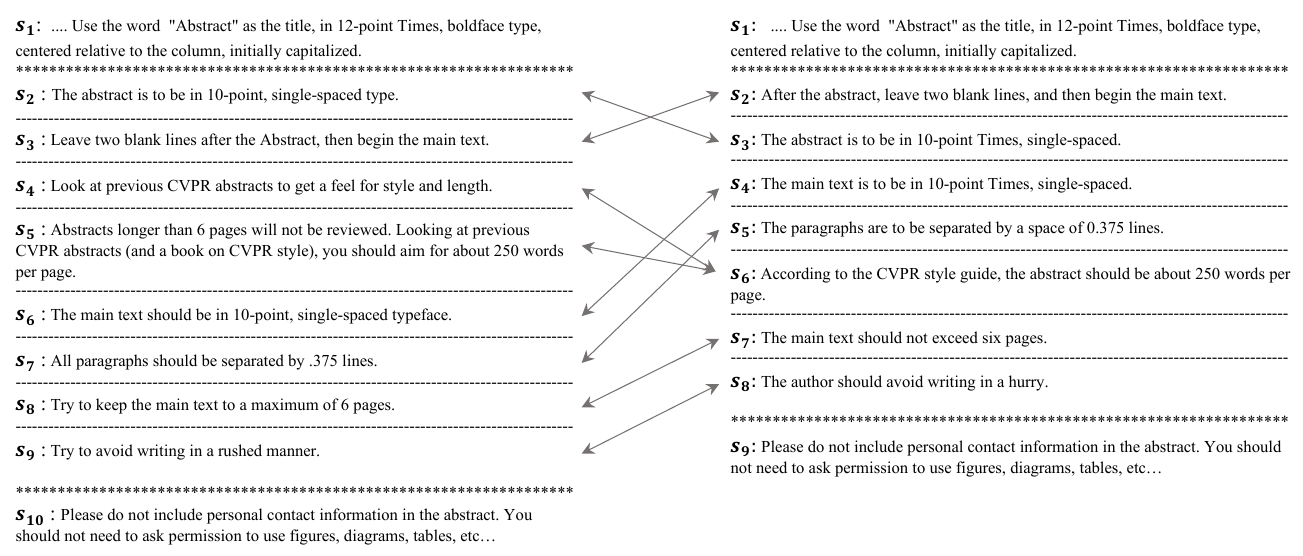}
\caption{
A case of paraphrasing involving sentence reordering and many-to-one sentence alignment is presented.
The original text exhibits changes in the order of $s_2$ and $s_3$, as well as the order of $s_4$--$s_5$ and $s_6$--$s_7$, after paraphrasing.
In particular, the text span $s_4$--$s_5$ in the original text aligns with $s_6$ in the paraphrased text.
}
\label{fig:para_case}
\end{figure*}

We show a case where paraphrasing text involves sentence reordering and many-to-one sentence alignment in Figure~\ref{fig:para_case}.
The detailed implementation for aligning paraphrased sentences with their reference original sentences is presented in Algorithm~\ref{alg:align}.
Our approach involves greedily traversing each paraphrased sentence and identifying a span of consecutive original sentences that share high semantic similarity.
If no suitable span is found, we fallback to finding the most semantically similar original sentence.
To determine a similarity threshold, we sample 1,000 instances where the number of sentences remains unchanged during paraphrasing, where most cases exhibit trivial one-on-one alignment upon manual inspection. 
We compute the semantic similarity between all pairs of paraphrases and original sentences. 
Pairs with matching indices (e.g, 0 v.s. 0) are considered matched while others are unmatched  (e.g, 0 v.s. 1 and 2).
The distribution of similarities is depicted in Figure~\ref{fig:para_bound}, suggesting that 0.75 is a promising threshold value.

\begin{figure}[t!]
\centering
\includegraphics[width=0.9\linewidth]{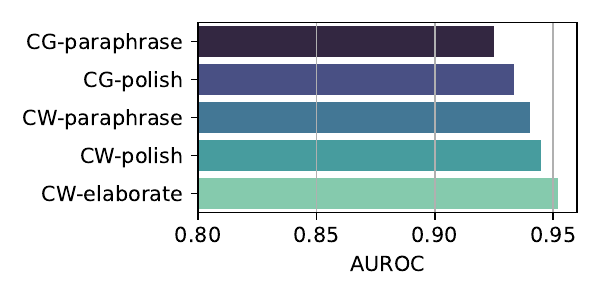}
\caption{
\label{fig:effects_prompts}
Effect of prompt types on detection performance (lexical regression). ``CG'' stands for ``context-agnostic'' and ``CW'' stands for context-aware. Prompt type instances can be referred to in Appendix~\ref{app:prompt}.
}
\end{figure}

\begin{algorithm}[t!]
\small
\textbf{Setup:} $n$: number of paraphrased sentences \\
\hspace*{2.7em} $m$: number of reference sentences \\
\hspace*{2.7em} $avg(L,i,j )$: calculate mean value of L[i,i+1,...j-1]  \\

\textbf{Input:} $mat$: Similarity matrix between paraphrased sentences \hspace*{4.7em}  and reference sentences, $R^{n\times m}$ \\
\hspace*{2.7em}  $\tau$: threshold
\begin{algorithmic}[1]
\STATE \COMMENT{initializing} 
\STATE $A \gets \emptyset $
\STATE $i\gets 0$  
\STATE \COMMENT{align each paraphrased sentence individually.}
\WHILE{$i < n$}
    \STATE \COMMENT{get the similarity of the most similar reference sentence}
    \STATE $V_{max} \gets mat[i].max()$

    \IF{$V_{max}\leq\tau$}
        \STATE \COMMENT{if the max similarity is less than or equal to the threshold, we just align the paraphrased sentence $i$ with the most similar reference.}
        \STATE $idx \gets argmax(mat[i])$
        \STATE $A.add((i,(idx,idx+1)))$
    \ELSE
        \STATE \COMMENT{If the maximum similarity exceeds the threshold, we align paraphrased sentence $i$ with the longest reference sentence span that has an average similarity greater than the threshold.}
        \STATE $W_{size}\gets m$
        \STATE $Flag \gets 0$
        \WHILE{$W_{size}\geq 1$}
            \STATE $j\gets 0$
            \WHILE{$j\leq m-W_{size}$}
                \STATE $V_{mean} \gets avg(mat[i],j,j+W_{size})$
                \IF{$V_{mean}>\tau$}
                    \STATE  $A.add((i,(j,j+W_{size})))$
                    \STATE $Flag \gets 1$
                    \STATE \textbf{break}
                \ENDIF
                \STATE $j \gets j+1$
            \ENDWHILE
            \IF{$Flag = 1$}
                \STATE \textbf{break}
            \ENDIF
            \STATE $W_{size} \gets W_{size}-1$
        \ENDWHILE
    \ENDIF
    \STATE $i\gets i+1$
\ENDWHILE
\RETURN $A$
\end{algorithmic}
\caption{Paraphrase Alignment}
\label{alg:align}
\end{algorithm}

\section{Data Statistics}
\label{app:stat}
The data statistics of PASTED dataset is presented in Table~\ref{tab:stat}.

\begin{table*}[t]
\setlength{\belowcaptionskip}{-0.3cm}
\centering
\small
\begin{tabular}{cccc}
    \toprule
     \textbf{Data Source} & \textbf{Human}& \textbf{Machine} & \textbf{All} \\
     \midrule
      \textbf{Original Texts} & 6,577/857/838   &  16,082/2,038/2,081  & 22,659/2,895/2,919 \\
      \midrule 
      \textbf{Context-agnostic Paraphrases} (ChatGPT) & 6,577/857/838  &  16,082/2,038/2,081  & 22,659/2,895/2,919 \\
      \midrule
      \textbf{Context-aware Paraphrases} (Dipper) & 6,235/830/796  &  14,477/1,937/1,868  & 20,712/2,767/2,664\\
      \midrule
      \textbf{All} & 20,712/2,767/2,664   &  46,641/6,013/6,030  & 66,030/8,557/8,502 \\
\bottomrule
\end{tabular}
\vspace{-0.2cm}
    \caption{In-distribution data statistics. The number of train, validation and test set is delimited by ``/''.}
    \label{tab:stat}
\end{table*}

\section{Data Samples}
\label{app:data_sample}

\begin{table*}[t!]
    \setlength{\belowcaptionskip}{-0.cm}
    \centering
    \small
    \renewcommand{\arraystretch}{1.2} 
    \begin{tabular}{lp{0.8\linewidth}} 
    \toprule
    
      \textbf{Raw Text 1} &  In great anticipation, we made our way from our hotel to the restaurant. We were there at 5:45 and had no problems being seated right away.
        \textcolor{rawcolor}{We had the 14 nam kao tod meat salad, garlic prawns, tom kha gai soup and crispy duck on drunken noodles. It was a lot of food, enough for a doggie bag for later too. To drink, Thai ice tea (creamy almost like an iced coffee) and a Thai Lime drink (refreshing but too sweet, ask for less sugar). Total came to around 75.00 for the two of us. Now, I may be crucified for this but, it wasn't knock my socks off amazing.
        }
    It was delicious, but I've been spoiled with Thai food here in Toronto. We go to Express Thai on Dufferin which is owned and operated by a Thai family, and I feel the flavours there are just as fresh and authentic as Lotus of Siam. 
    \\
     \midrule
      \textbf{CG-Para.  1} &  In great anticipation, we made our way from our hotel to the restaurant. We were there at 5:45 and had no problems being seated right away.
      \textcolor{paracolor}{ We ordered various dishes including the 14 nam kao tod meat salad, garlic prawns, tom kha gai soup, and crispy duck on drunken noodles. The portion sizes were generous, so we even had leftovers for later. For our beverages, we had Thai ice tea, which had a creamy texture similar to iced coffee, and a Thai Lime drink, which was refreshing but slightly too sweet. The bill totaled around \$75. While some might strongly disagree, I personally didn't find the food to be exceptionally outstanding.}
     It was delicious, but I've been spoiled with Thai food here in Toronto. We go to Express Thai on Dufferin which is owned and operated by a Thai family, and I feel the flavours there are just as fresh and authentic as Lotus of Siam. \\
      \midrule 
      \textbf{CA-Para.  1} &  In great anticipation, we made our way from our hotel to the restaurant. We were there at 5:45 and had no problems being seated right away.
      \textcolor{paracolor}{ We ordered the nam kao tod, a meat salad, garlic shrimp, tom yam goong, and duck with drunken noodles. We also had a Thai iced tea and a lime juice, both of which were very sweet. I’m afraid I’m going to be criticized for saying this, but the food wasn’t that great. It was good, but not great. The bill came to about \$70 for the two of us.
    }
     It was delicious, but I've been spoiled with Thai food here in Toronto. We go to Express Thai on Dufferin which is owned and operated by a Thai family, and I feel the flavours there are just as fresh and authentic as Lotus of Siam. \\

    \midrule
      \textbf{Raw Text 2} & My friend and I thought it would be fun to try Pub 1842 for a light lunch, considering all we do is eat when we go to Vegas.
        \textcolor{rawcolor}{I can't lie, my friend has an obsession with buffets (MGM buffet is the best) but I was trying to get her to eat something different and on the fancy side. In the middle of the MGM casino, you'll find this restaurant with a great atmosphere. They have a bar area as well, which is perfect for a drink while you're waiting for your table. We sat outside on the patio, which was nice because it was bright and sunny outside. I had the lunch special which consisted of a mixed green salad with candied walnuts, gorgonzola cheese, and dried cranberries. The salad was huge and I was only able to eat half of it. The dressing was on the sweet side, I like mine to be a bit more tangy. The 1842 Burger is the house specialty.
        }
     I couldn't resist because it had a fried egg on it, and anything with an egg is always a good thing! The sweet potato fries were great! Crispy and perfect on the side. I had the funnel cake for dessert, which was just okay. Not the best funnel cake I've had.

        \\
     \midrule
      \textbf{CG-Para.  2} & My friend and I thought it would be fun to try Pub 1842 for a light lunch, considering all we do is eat when we go to Vegas.
      \textcolor{paracolor}{ 
    Honestly, my friend is really obsessed with buffets (specifically the MGM buffet), but I wanted to convince her to try something different and more upscale. Right in the middle of the MGM casino, there's this restaurant with a fantastic atmosphere. They even have a bar area, which is perfect for enjoying a drink while waiting for your table. We decided to sit outside on the patio, which was great because it was sunny and bright outdoors. For my meal, I opted for the lunch special, which included a generous mixed green salad topped with candied walnuts, gorgonzola cheese, and dried cranberries. The salad was enormous, and I could only manage to eat half of it. The dressing was on the sweeter side, although I prefer mine to have a tangy kick. The house specialty here is the 1842 Burger.
    }
    I couldn't resist because it had a fried egg on it, and anything with an egg is always a good thing! The sweet potato fries were great! Crispy and perfect on the side. I had the funnel cake for dessert, which was just okay. Not the best funnel cake I've had.

    \\
      \midrule 
      \textbf{CA-Para.  2} &  My friend and I thought it would be fun to try Pub 1842 for a light lunch, considering all we do is eat when we go to Vegas.
      \textcolor{paracolor}{  In the middle of the MGM Grand Casino, this restaurant has a great atmosphere and a bar that is perfect for a drink while you wait for a table. I won’t lie, my friend is obsessed with buffets (the MGM Grand buffet is the best), but I wanted to try something different and a little more upscale. I ordered the salad special, which was a mixed green salad with candied walnuts, gorgonzola cheese, and dried cranberries. It was so big I could only eat half of it. We sat outside on the patio, which was nice because it was sunny. The 1842 burger is the specialty. The salad dressing was a little too sweet for my taste.
     }
   I couldn't resist because it had a fried egg on it, and anything with an egg is always a good thing! The sweet potato fries were great! Crispy and perfect on the side. I had the funnel cake for dessert, which was just okay. Not the best funnel cake I've had.
\\

    \bottomrule
    \end{tabular}
    \caption{
    \label{tab:sample1}
     Data samples in PASTED, where ``CG-Para.'' stands for context-agnostic paraphrasing and ``CW-Para.'' denotes context-aware paraphrasing.
    }
\end{table*}

\begin{table*}[t!]
    \setlength{\belowcaptionskip}{-0.cm}
    \centering
    \small
    \renewcommand{\arraystretch}{1.2} 
    \begin{tabular}{lp{0.8\linewidth}} 
    \toprule
    
      \textbf{Raw Text 3} & Make sure to read the manual and allow yourself a few hours to explore all the options. Once you're comfortable with one, train your fingers to type or swipe quicker.
        \textcolor{rawcolor}{This is a must if you're a power user. You'll get the most out of your smartphone by keeping it updated to the latest version of the software. You may also notice an app or two that isn't working properly. Simply uninstall it.
        }
   Also keep a close eye on your storage space. If you're running out, you may need to delete some pictures, videos and files. Your smartphone is so much more than a phone.\\
     \midrule
      \textbf{CG-Para.  3} &  Make sure to read the manual and allow yourself a few hours to explore all the options. Once you're comfortable with one, train your fingers to type or swipe quicker.
      \textcolor{paracolor}{ 
      If you regularly use your smartphone, it is essential to keep it updated with the latest software version for optimal performance. Additionally, you might come across a few malfunctioning apps, in which case, you can easily remove them.
      }
      If you regularly use your smartphone, it is Also keep a close eye on your storage space. If you're running out, you may need to delete some pictures, videos and files. Your smartphone is so much more than a phone.\\
      \midrule 
      \textbf{CA-Para.  3} &  Make sure to read the manual and allow yourself a few hours to explore all the options. Once you're comfortable with one, train your fingers to type or swipe quicker.
      \textcolor{paracolor}{  You'll also notice that some of the apps aren't working properly. If you're a power user, you'll want to keep your phone's operating system up to date. If an app isn't working, just uninstall it.
    }
     Also keep a close eye on your storage space. If you're running out, you may need to delete some pictures, videos and files. Your smartphone is so much more than a phone. \\

     \midrule
     
     \textbf{Raw Text 4} & Atticus does not want Jem and Scout to be present at Tom Robinson's trial. No seat is available on the main floor, so by invitation of Rev.
    \textcolor{rawcolor}{Sykes, Jem, Scout, and Dill watch from the colored balcony. Atticus establishes that the accusers - Mayella and her father, Bob Ewell, the town drunk - are lying. It also becomes clear that the friendless Mayella made sexual advances toward Tom, and that her father caught her and beat her. Despite significant evidence of Tom's innocence, the jury convicts him.
    }
        Jem's faith in justice becomes badly shaken, as is Atticus', when the hapless Tom is shot and killed while trying to escape from prison.
\\
  
     \midrule
      \textbf{CG-Para.  4} &  Atticus does not want Jem and Scout to be present at Tom Robinson's trial. No seat is available on the main floor, so by invitation of Rev.
      \textcolor{paracolor}{ 
     Sykes, Jem, Scout, and Dill observe the events from the balcony designated for the colored people. Atticus proves that Mayella and her father, Bob Ewell, who is known as the town drunk, are not telling the truth. It also becomes evident that Mayella, who has no friends, made sexual advances towards Tom, and her father discovered it and violently beat her. Although there is substantial evidence to prove Tom's innocence, the jury unfairly finds him guilty.
      }
      Jem's faith in justice becomes badly shaken, as is Atticus', when the hapless Tom is shot and killed while trying to escape from prison.
\\
      \midrule 
      \textbf{CA-Para.  4} &  Atticus does not want Jem and Scout to be present at Tom Robinson's trial. No seat is available on the main floor, so by invitation of Rev.
      \textcolor{paracolor}{  Sykes, Jem, Scout, and Dill are allowed to sit in the balcony. Atticus proves that the accusers, Mayella and her father Bob Ewell, are lying. It also becomes clear that Mayella, who has no friends, had been trying to seduce Tom, but her father had caught her and beaten her. Tom Robinson is convicted, despite the overwhelming evidence of his innocence.
    }
     Jem's faith in justice becomes badly shaken, as is Atticus', when the hapless Tom is shot and killed while trying to escape from prison.
     \\

    \bottomrule
    \end{tabular}
    \caption{
    \label{tab:sample2}
     Data samples in PASTED, where ``CG-Para.'' stands for context-agnostic paraphrasing and ``CW-Para.'' denotes context-aware paraphrasing.
    }
\end{table*}

We present several data samples in Table~\ref{tab:sample1} and Table~\ref{tab:sample2}.

\section{Prompt Design}
\label{app:prompt}
\begin{figure*}[t!]
\centering
\includegraphics[width=0.99\linewidth]{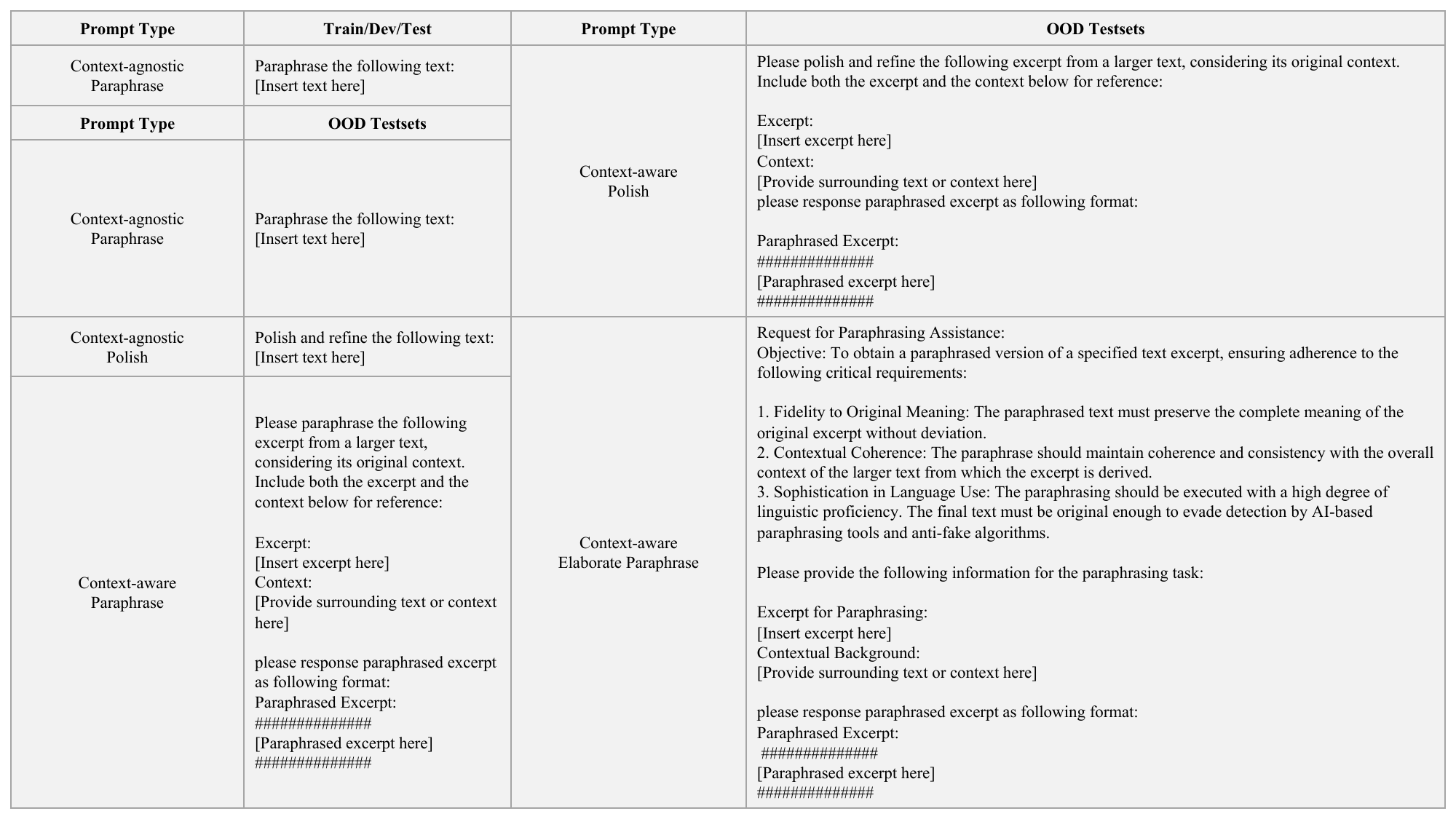}
\caption{
\label{fig:prompt}
Case illustration for 5 prompt types used in this paper. ``Context-agnostic Paraphrase'' is used for constructing the in-distribution train and test set, while the other prompts types are used for constructing the generalization testset.
}
\end{figure*}

The prompt templates used for constructing the data set is shown in Figure~\ref{fig:prompt}.

\section{Word Cloud of Paraphrases}
\begin{figure*}[t!]
\centering
\includegraphics[width=1\linewidth]{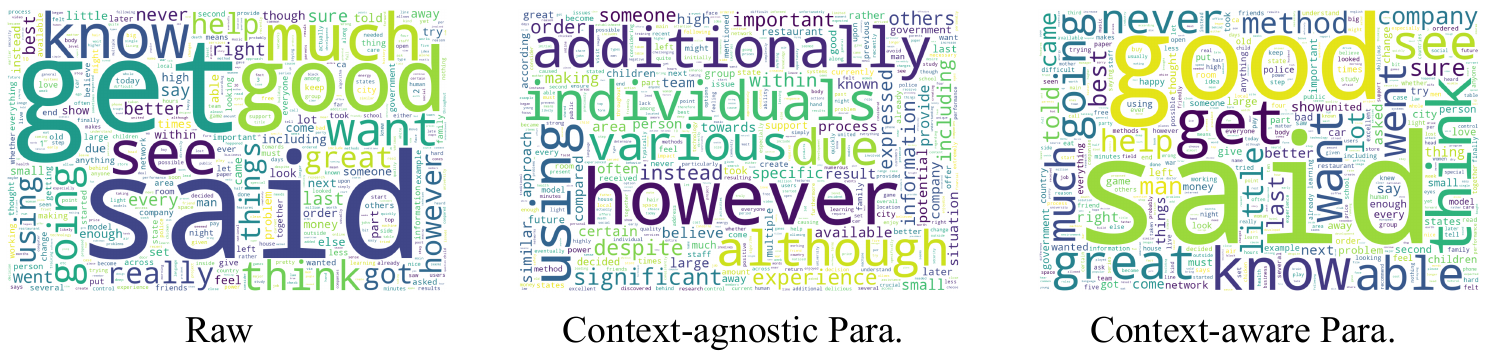}
\caption{
\label{fig:word_cloud}
Word clouds of original texts, context-agnostic paraphrases and context-aware paraphrases.
}
\end{figure*}

The word cloud of word distribution for original texts, context-agnostic paraphrases, and context-aware paraphrases are shown in Figure~\ref{fig:word_cloud}.
\label{app:word_cloud}

\section{Case Study}
\label{app:case_study}
\begin{figure*}[t]
\centering
\includegraphics[width=0.99\linewidth]{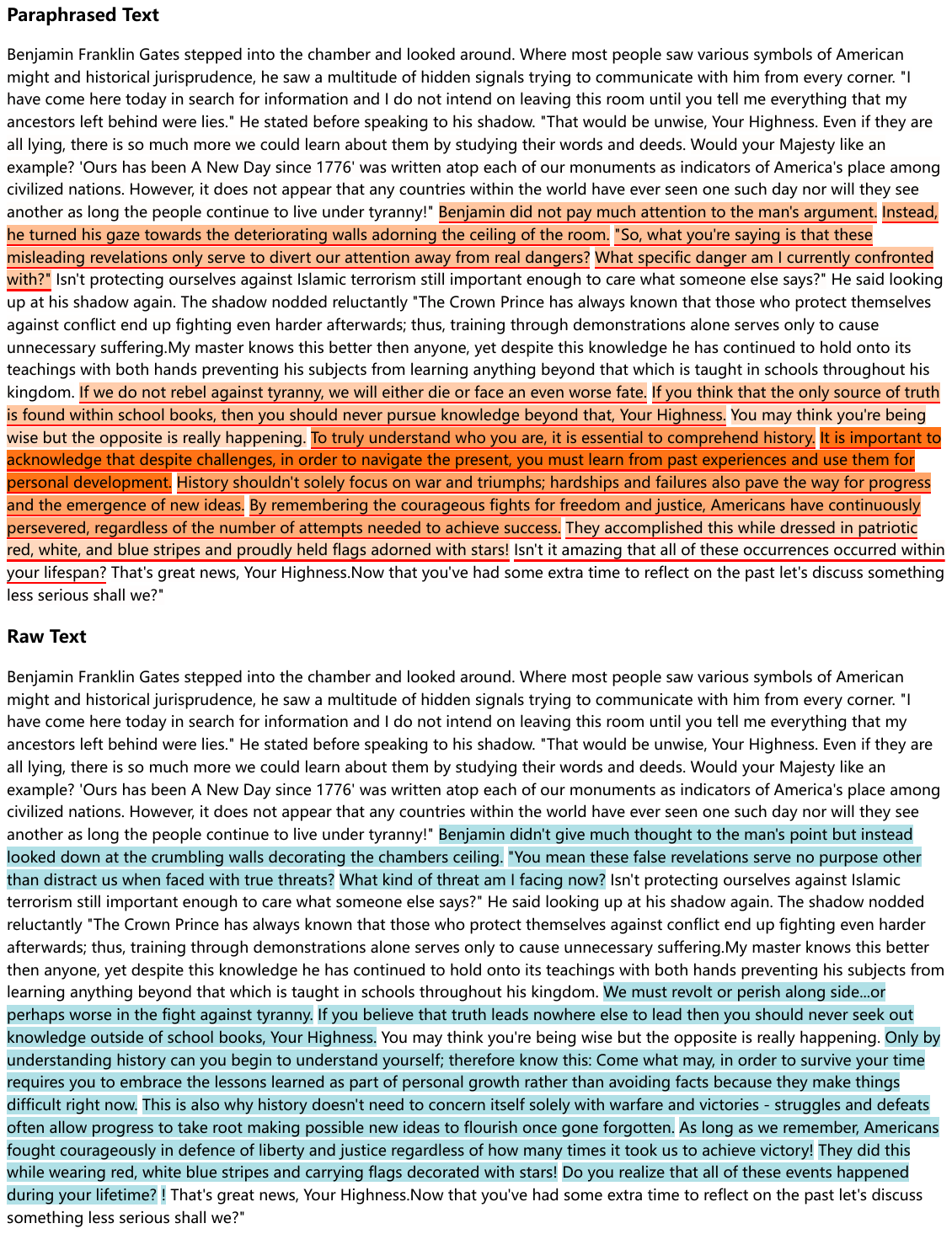}
\caption{
\label{fig:case_study1}
A case study of paraphrased text span detection. The upper part presents the paraphrased text while the lower part denotes the original text.
The red underlined text represents the paraphrased text spans, and the orange background indicates model predictions. Darker colors indicate a higher degree of paraphrasing.
The text in the blue background represents the original text span before paraphrasing.
}
\end{figure*}

\begin{figure*}[t]
\centering
\includegraphics[width=0.99\linewidth]{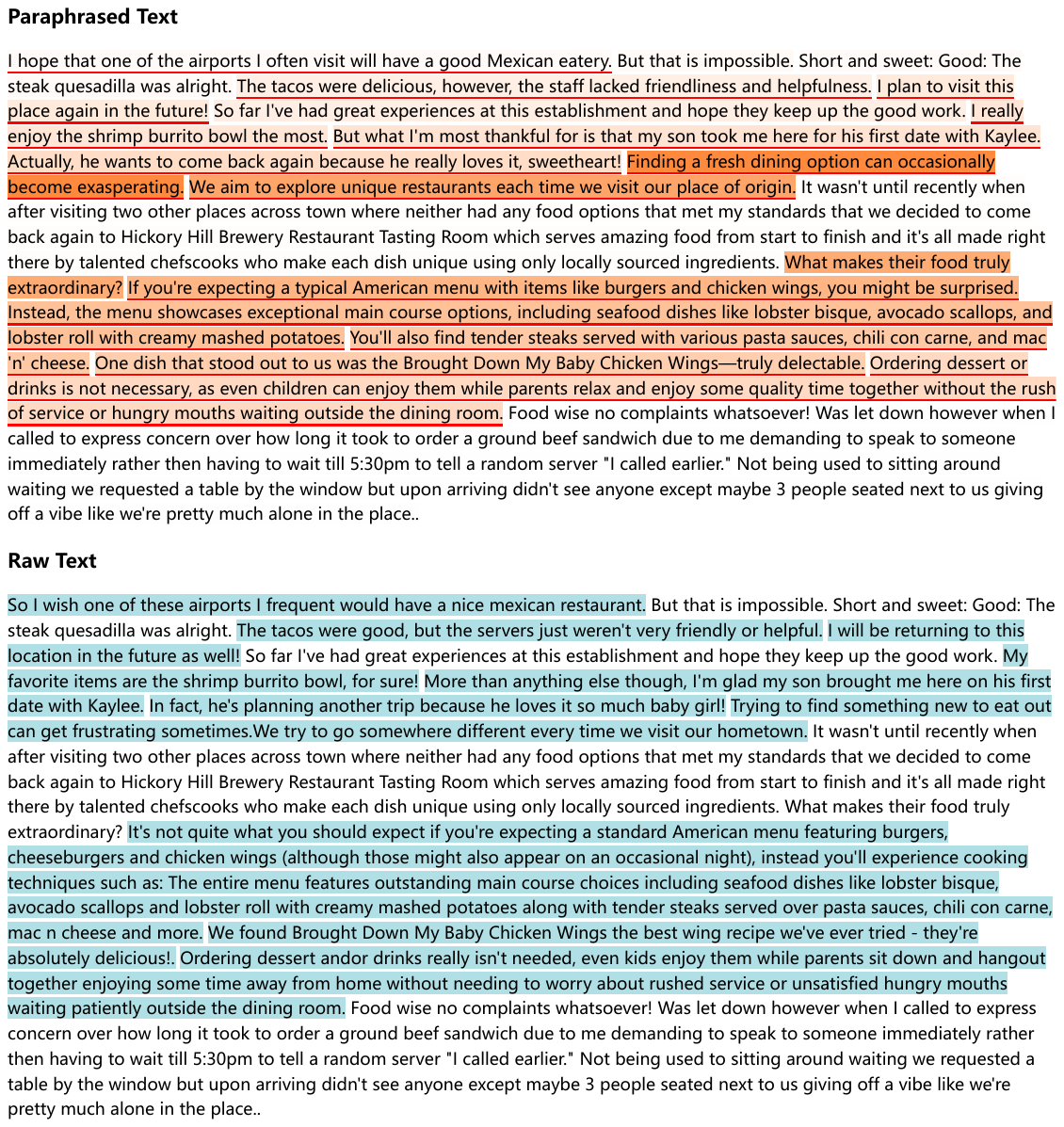}
\caption{
\label{fig:case_study2}
A case study of paraphrased text span detection. The upper part presents the paraphrased text while the lower part denotes the original text.
The red underlined text represents the paraphrased text spans, and the orange background indicates model predictions. Darker colors indicate a higher degree of paraphrasing.
The text in the blue background represents the original text span before paraphrasing.
}
\end{figure*}

\begin{figure*}[t]
\centering
\includegraphics[width=0.99\linewidth]{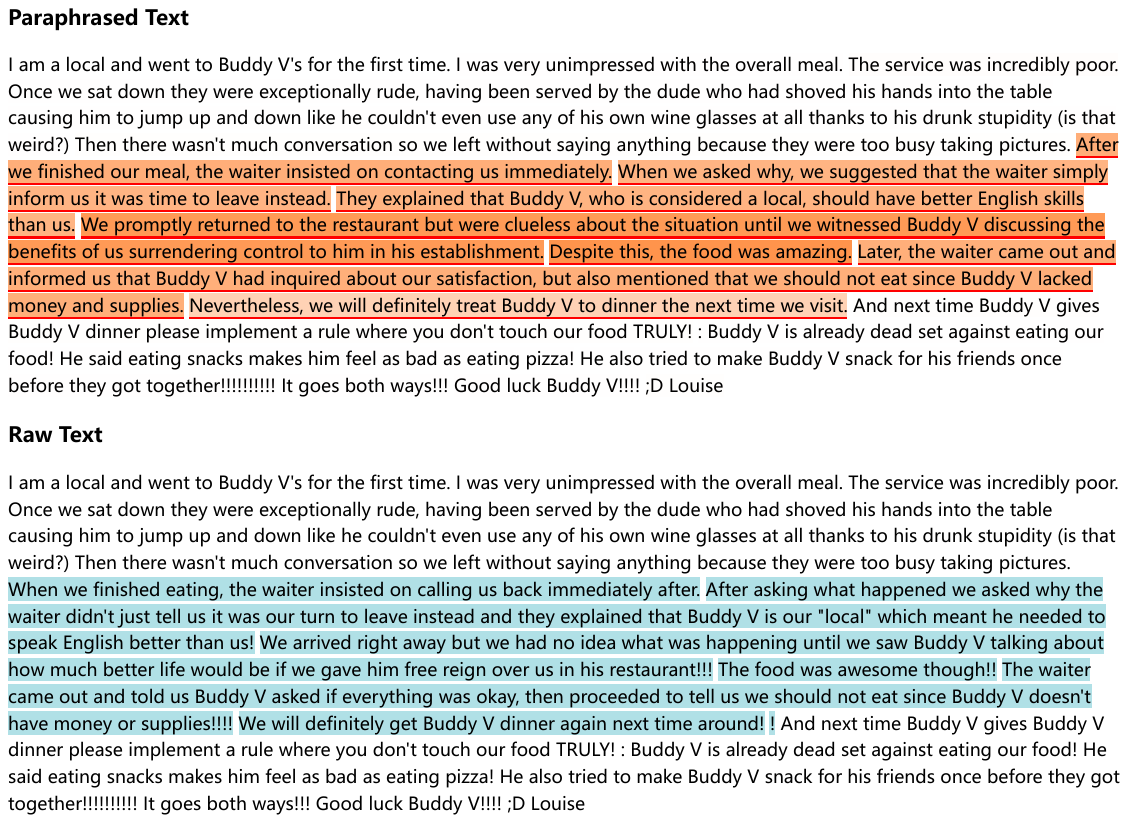}
\caption{
\label{fig:case_study3}
A case study of paraphrased text span detection. The upper part presents the paraphrased text while the lower part denotes the original text.
The red underlined text represents the paraphrased text spans, and the orange background indicates model predictions. Darker colors indicate a higher degree of paraphrasing.
The text in the blue background represents the original text span before paraphrasing.
}
\end{figure*}

We present several cases of paraphrased text span detection (aggregate regression model) in Figure~\ref{fig:case_study1}, Figure~\ref{fig:case_study2} and Figure~\ref{fig:case_study3}.

\section{Effect of Prompts}
\label{app:effects_prompts}
The effect of prompts for constructing paraphrases is shown in Figure~\ref{fig:effects_prompts}, where ``CG'' and ``CW'' are abbreviations for ``context-agnostic'' and ``context-aware'' respectively.


\section{Constructing Texts with Minor Modification}
\label{app:robustness}
For sentence reordering, we set the lexical control of Dipper to the lowest value (i.e., 20) and the order control value to the max (i.e., 100)\footnote{Please refer to https://github.com/martiansideofthemoon/ai-detection-paraphrases for details.}.
For word replacement, we randomly mask 10\% of the tokens and use T5-3B model to fill in these blanks.
We apply both methods on the in-distribution testset.
To ensure minimal modifications, we calculate a similarity score between the original and paraphrased text. We then establish a threshold to exclude paraphrases that significantly differ from the original texts.
We only retain paraphrases with BLEU scores above 70 when compared with their respective original texts.
In this way, we construct 273 and 1,310 instances for sentence reordering and word replacement, respectively.
We show several cases in Table~\ref{tab:minor_case}.

\begin{table*}[t!]
    \setlength{\belowcaptionskip}{-0.cm}
    \centering
    \small
    \renewcommand{\arraystretch}{1.2} 
    \begin{tabular}{lp{0.8\linewidth}} 
    \toprule
    \multicolumn{2}{c}{\textbf{Sentence Reordering}} \\
    \midrule
      \textbf{Raw Text 1} & It felt good to know that his efforts were appreciated. At the end of the week, Paul was exhausted but satisfied.  \\
      \midrule
      \textbf{Paraphrase 1} & At the end of the week, Paul was exhausted, but satisfied. It felt good to know that his efforts were appreciated. \\
    \midrule
      \textbf{Raw Text 2} & Its happened a few times, like some gas leak in some town and it instructed for people to be indoors and shut all the windows. Signal intrusion has happened before as someone already posted. But encryption makes that all but impossible today. You would either need some serious ability to break encryption or infiltrate multiple tv providers to pull it off.  \\
      \midrule
      \textbf{Paraphrase 2} & Signal intrusion has happened before as someone already posted. But encryption makes that all but impossible today. You would either need some serious ability to break encryption or infiltrate multiple TV broadcasters to pull it off. It's happened a few times, like some gas leak in some town and it instructed for people to be indoors and shut all the windows. \\
    \midrule
      \textbf{Raw Text 3} & Keep the toes of your forward foot pointing upwards. To choose which leg to bend, try both out and see which one feels most comfortable to you. Determine how far apart the legs should be when sitting on a chair or stool: If they are together, then position yourself so that your heels touch. \\
      \midrule
      \textbf{Paraphrase 3} & Determine how far apart the legs should be when sitting on a chair or a stool: if they are together, then position yourself so that your heels touch. Keep the toes of the front foot pointing upwards. To choose which leg to bend, try both legs and see which one is more comfortable to you. \\
    \midrule
    \multicolumn{2}{c}{\textbf{Word Replacement}} \\
    \midrule
      \textbf{Raw Text 4} & Among those arrested were six suspects in Italy, four in Britain, and three in Norway. Police say some of the suspects may have travelled to Syria or Iraq. Italy's Ansa Six suspected Ansar al-Sharia fighters were arrested in Italy, Britain, Pakistan, Norway, Scotland and Germany, bringing the total to 18. Police say they have arrested a further 22 people on suspicion of involvement in terrorism.  \\
      \midrule
      \textbf{Paraphrase 4} & Among those arrested were six suspects in the United States, four in Britain, and three in Norway. Police say some of the suspects may have travelled to Syria or Iraq. Italy's Ansa Six suspected Jaish al-Sharia fighters were arrested in Italy, Britain, Pakistan, Norway, France and Germany, bringing and the total to 181. Police say they have arrested a further 22 people on suspicion of involvement in terrorism. \\ 
    \midrule
      \textbf{Raw Text 5} & Soviet Union. Russia was a state of the Soviet Union. It technically took the Germans a while to reach Russia after they invaded.  \\
      \midrule
      \textbf{Paraphrase 5} & Soviet Union. Russia was a state of the Soviet Union. It also took the Germans a while to conquer Russia after they invaded. \\ 
    \bottomrule
    \end{tabular}
    \caption{
    \label{tab:minor_case}
    Case illustration of two types of minor modifications: sentence reordering and word replacement.
    }
    
\end{table*}

\end{document}